\documentclass[runningheads]{llncs}

\usepackage{eccv}

\usepackage{eccvabbrv}

\usepackage{graphicx}
\usepackage{booktabs}

\usepackage[accsupp]{axessibility}  %

\usepackage[pagebackref,breaklinks,colorlinks,citecolor=eccvblue]{hyperref}
\usepackage{orcidlink}

\usepackage[dvipsnames]{xcolor}

\newcommand{\todo}[1]{{\color{red}#1}}

\newcommand{\xpm}[1]{{\tiny$\pm#1$}}

\makeatletter
\renewcommand{\paragraph}{%
\@startsection{paragraph}{4}%
{\z@}{0.5em}{-1em}%
{\normalfont\normalsize\bfseries}%
}
\makeatother

\definecolor{blue}{RGB}{20, 105, 181}
\definecolor{yellow}{RGB}{249, 199, 79}
\definecolor{green}{RGB}{104, 200, 59}
\definecolor{orange}{RGB}{249, 102, 44}

\usepackage{gensymb}
\usepackage{amsmath}
\usepackage[capitalize]{cleveref}
\usepackage{graphicx}
\usepackage{xspace}
\usepackage{multirow}
\usepackage{makecell}

\usepackage[capitalize]{cleveref}
\Crefname{section}{Section}{Sections}
\crefname{section}{Sec.}{Secs.}
\Crefname{figure}{Figure}{Figures}
\crefname{figure}{Fig.}{Figs.}
\Crefname{table}{Table}{Tables}
\crefname{table}{Tab.}{Tabs.}
\Crefname{Equation}{Equation}{Equations}
\crefname{equation}{Eq.}{Eqs.}
\usepackage{url}            %
\usepackage{amsfonts}       %
\usepackage{nicefrac}       %
\usepackage{microtype}      %
\usepackage[dvipsnames]{xcolor}         %

\newif\ifdrafting
\draftingtrue %
\ifdrafting
    \newcommand{\diff}[2]{{\color{red}\sout{#1}} {\color{green}#2}}
    \newcommand{\missing}[1]{{\color{orange}{#1}}}
    
    \newcommand{\vj}[1]{{\color{magenta}{\textbf{Varun: } #1}}}
    \newcommand{\ar}[1]{{\color{green}{\textbf{AR: } #1}}}
\else
    \newcommand{\todo}[1]{}
    \newcommand{\diff}[1]{}
    \newcommand{\missing}[1]{}
    \newcommand{\vj}[1]{}
    \newcommand{\ar}[1]{}
\fi

\usepackage{algorithm}
\usepackage{algpseudocode}
\usepackage{wrapfig}

\def\task{3D Congealing\xspace}

\begin{document}

\title{\task: 3D-Aware \\Image Alignment in the Wild} 

\author{Yunzhi Zhang\inst{1} \and
Zizhang Li\inst{1} \and
Amit Raj\inst{2}\and Andreas Engelhardt\inst{3}\and Yuanzhen Li\inst{2}
\and Tingbo Hou\inst{2} \and Jiajun Wu\inst{1}
\and Varun Jampani\inst{4}}

\authorrunning{Y.~Zhang et al.}
\institute{Stanford University \and
Google Research\and University of Tübingen
\and Stability AI}

\maketitle

\begin{abstract}
We propose \task, a novel problem of 3D-aware alignment for 2D images capturing semantically similar objects. Given a collection of unlabeled Internet images, our goal is to associate the shared semantic parts from the inputs and aggregate the knowledge from 2D images to a shared 3D canonical space. 
We introduce a general framework that tackles the task without assuming shape templates, poses, or any camera parameters. At its core is a canonical 3D representation that encapsulates geometric and semantic information. The framework optimizes for the canonical representation together with the pose for each input image, and a per-image coordinate map that warps 2D pixel coordinates to the 3D canonical frame to account for the shape matching.
The optimization procedure fuses prior knowledge from a pre-trained image generative model and semantic information from input images.
The former provides strong knowledge guidance for this under-constraint task, while the latter provides the necessary information to mitigate the training data bias from the pre-trained model. 
Our framework can be used for various tasks such as 
pose estimation and image editing, achieving strong results on real-world image datasets under challenging illumination conditions and on in-the-wild online image collections. 
Project page at \footnotesize{\url{https://ai.stanford.edu/~yzzhang/projects/3d-congealing/}}. 

\end{abstract}
\begin{figure}[t]
    \centering
\includegraphics[width=\linewidth]{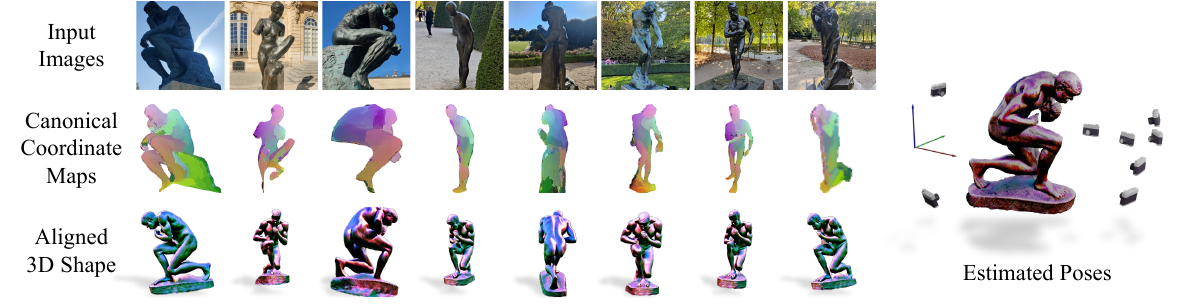}
\captionof{figure}{
Objects with different shapes and appearances, such as these sculptures, may share similar semantic parts and a similar geometric structure. We study \task, inferring and aligning such a shared structure from an unlabeled image collection. %
Such alignment can be used for tasks such as pose estimation and image editing. See \cref{app:rodin_full} for full results.
\label{fig:teaser}
}
\vspace{-1em}
\end{figure}

\section{Introduction}
\label{sect:introduction}

We propose the task of \emph{\task}, where the goal is to align a collection of images containing semantically similar objects into a shared 3D space. Specifically, we aim to obtain a canonical 3D representation together with the pose and a dense map of 2D-3D correspondence for each image in the collection. The input images may contain object instances belonging to a similar category with varying shapes and textures, and are captured under distinct camera viewpoints and illumination conditions, which all contribute to the pixel-level difference as shown in \Cref{fig:teaser}.
Despite such inter-image differences, humans excel at aligning such images with one another in a geometrically and semantically consistent manner based on their 3D-aware understanding. %

Obtaining a canonical 3D representation and grounding input images to the 3D canonical space enable several downstream tasks, such as 6-DoF object pose estimation, pose-aware image filtering, and image editing.
Unlike the task of 2D congealing~\cite{ofri2023neural,peebles2022gan,gupta2023asic}, where the aim is to align the 2D pixels across the images, \task requires aggregating the information from the image collection altogether and forming the association among images in 3D. 
The task is also closely related to 3D reconstruction from multiview images, with a key distinction in the problem setting, as inputs here do not necessarily contain identical objects but rather semantically similar ones. Such a difference opens up the possibility of image alignment from readily available image collections on the Internet, \eg, online search results, landmark images, and personal photo collections.

\task represents a challenging problem, particularly for arbitrary images without camera pose or lighting annotations, even when the input images contain identical objects~\cite{lin2021barf,chen2023local,boss2022samurai,wang2021nerf}, because the solutions for pose and shape are generally entangled. 
On the one hand, the definition of poses is specific to the coordinate frame of the shape; on the other hand, the shape optimization is typically guided by the pixel-wise supervision of images under the estimated poses. 
To overcome the ambiguity in jointly estimating poses and shapes, prior works mostly start from noisy pose initializations~\cite{lin2021barf}, data-specific initial pose distributions~\cite{wang2021nerf,meng2021gnerf}, or rough pose annotations such as pose quadrants~\cite{boss2022samurai}. They then perform joint optimization for a 3D representation using an objective of reconstructing input image pixels~\cite{wang2021nerf,lin2021barf,boss2022samurai} or distribution matching~\cite{meng2021gnerf}. 

In this work, instead of relying on initial poses as starting points for shape reconstruction, we propose to tackle the joint optimization problem from a different perspective.
We first obtain a plausible 3D shape that is compliant with the input image observations using pre-trained generative models, and then use semantic-aware visual features, \eg, pre-trained features from DINO~\cite{caron2021emerging,oquab2023dinov2} and Stable-Diffusion~\cite{rombach2022high}, to register input images to the 3D shape. 
Compared to photometric reconstruction losses, these features are more tolerant of variance in object identities among image inputs. 

We make deliberate design choices to instantiate such a framework that fuses the knowledge from pre-trained text-to-image (T2I) generative models with real image inputs.
First, to utilize the prior knowledge from generative models, we opt to apply a T2I personalization method, Textual Inversion~\cite{gal2022image}, which aims to find the most suitable text embedding to reconstruct the input images via the pre-trained model. Furthermore, a semantic-aware distance is proposed to mitigate the appearance discrepancy between the rendered image and the input photo collection. Finally, a canonical coordinate mapping is learned to find the correspondence between 3D canonical representation and 2D input images.

To prove the effectiveness of the proposed framework, 
we compare the proposed method against several baselines on the task of pose estimation on a dataset with varying illuminations and show that our method surpasses all the baselines significantly. We also demonstrate several applications of the proposed method, including image editing and object alignment on web image data.

In summary, our contributions are:
\vspace{-2mm}
\begin{enumerate}
    \item We propose a novel task of \task that involves aligning images of semantically similar objects in a shared 3D space.
    \item We develop a framework tackling the proposed task and demonstrate several applications using the obtained 2D-3D correspondence, such as pose estimation and image editing.
    \item We show the effectiveness and applicability of the proposed method on a diverse range of in-the-wild Internet images. 
\end{enumerate}

\section{Related Works}

\paragraph{Image Congealing.}
Image congealing methods~\cite{huang2012learning,huang2007unsupervised,learned2005data,miller2000learning} aim to align a collection of input images based on the semantic similarity of parts. 
To tackle this task, Neural Congealing~\cite{ofri2023neural} proposes to use neural atlases, which are 2D feature grids, to capture common semantic features from input images and recover a dense mapping between each input image and the learned neural atlases. 
GANgealing~\cite{peebles2022gan} proposes to use a spatial transformer to map a randomly generated image from a GAN~\cite{goodfellow2020generative} to a jointly aligned space. These 2D-warping-based methods are typically applied to source and target image pairs with no or small camera rotation, and work best on in-plane transformation, while our proposed framework handles a \emph{larger variation of viewpoints} due to 3D reasoning. 

\paragraph{Object Pose Estimation.}
Object pose estimation aims to estimate the pose of an object instance with respect to the coordinate frame of its 3D shape. Classical methods for pose estimation recover poses from multi-view images using pixel- or feature-level matching to find the alignment between different images~\cite{schonberger2016structure}. These methods are less suitable in the in-the-wild setting due to the increasing 
variance among images compared to multi-view captures. 
Recent methods proposed to tackle this task by training a network with pose annotations as supervision~\cite{zhang2022relpose,lin2023relpose++,wang2023posediffusion}, but it remains challenging for these methods to generalize beyond the training distribution. Another class of methods propose to use an analysis-by-synthesis framework to estimate pose given category-specific templates~\cite{chen2020category} or a pre-trained 3D representation~\cite{yen2021inerf}; these assumptions make it challenging to apply these methods to generic objects in the real world.
ID-Pose~\cite{cheng2023id} leverages Zero-1-to-3~\cite{liu2023zero}, a view synthesis model, and optimizes for the relative pose given a source and a target image. Goodwin~\etal~\cite{goodwin2022zero} use pre-trained self-supervised features for matching, instead of doing it at the pixel level, but require both RGB and depth inputs; in contrast, we assume access to only RGB images. 

\paragraph{Shape Reconstruction from Image Collections.}
Neural rendering approaches~\cite{mildenhall2021nerf,wang2021neus,yariv2021volume} use images with known poses to reconstruct the 3D shape and appearance from a collection of multiview images. The assumptions of known poses %
and consistent illumination prevent these methods from being applied in the wild. Several works have extended these approaches to relax the pose assumption, proposing to handle noisy or unknown camera poses of input images through joint optimization of poses and 3D representation~\cite{wang2021nerf,lin2021barf,chen2023local}.  
SAMURAI~\cite{boss2022samurai} also proposes to extend the NeRF representation to handle scenes under various illuminations, provided with coarse initial poses in the form of pose quadrant annotations.
Unlike these methods, we do not have any assumption on the camera pose of input images, and handle image inputs with variations in illumination conditions. Moreover, %
our framework allows for aligning \emph{multiple objects} of the same category with geometric variations into a common coordinate frame.

\paragraph{3D Distillation from 2D Diffusion Models.}
Recently, text-to-image diffusion models have shown great advancement in 2D image generation, and DreamFusion~\cite{poole2022dreamfusion} has proposed to distill the prior on 2D images from pre-trained text-to-image models to obtain text-conditioned 3D representations. Other methods have extended this idea to optimize for 3D assets conditioned on image collections~\cite{raj2023dreambooth3d}. 
DreamFusion uses a full 3D representation, while Zero-1-to-3~\cite{liu2023zero} extracts the 3D knowledge using view synthesis tasks. Specifically, it finetunes Stable-Diffusion~\cite{rombach2022high} on synthetic data from Objaverse~\cite{deitke2023objaverse} to generate a novel view conditioned on an image and a relative pose. The fine-tuned model can be further combined with gradients proposed in DreamFusion to obtain full 3D assets. MVDream~\cite{shi2023mvdream} extends the idea of fine-tuning with view synthesis tasks, but proposes to output 4 views at once and has shown better view consistency in the final 3D asset outputs. DreamBooth3D~\cite{raj2023dreambooth3d} proposed to utilize fine-tuned diffusion model~\cite{ruiz2023dreambooth} for the image-conditioned 3D reconstruction task. Their goal is to faithfully reconstruct the 3D shape; therefore, the model requires a multi-stage training that involves applying diffusion model guidance and photometric-reconstruction-based updates. These works provide a viable solution for 3D reconstruction from image collections, but they do not \emph{ground the input images} to the 3D space as in ours.

\section{Method}
\label{sect:method}
\begin{figure}[t]
    \centering
    \includegraphics[width=\linewidth]{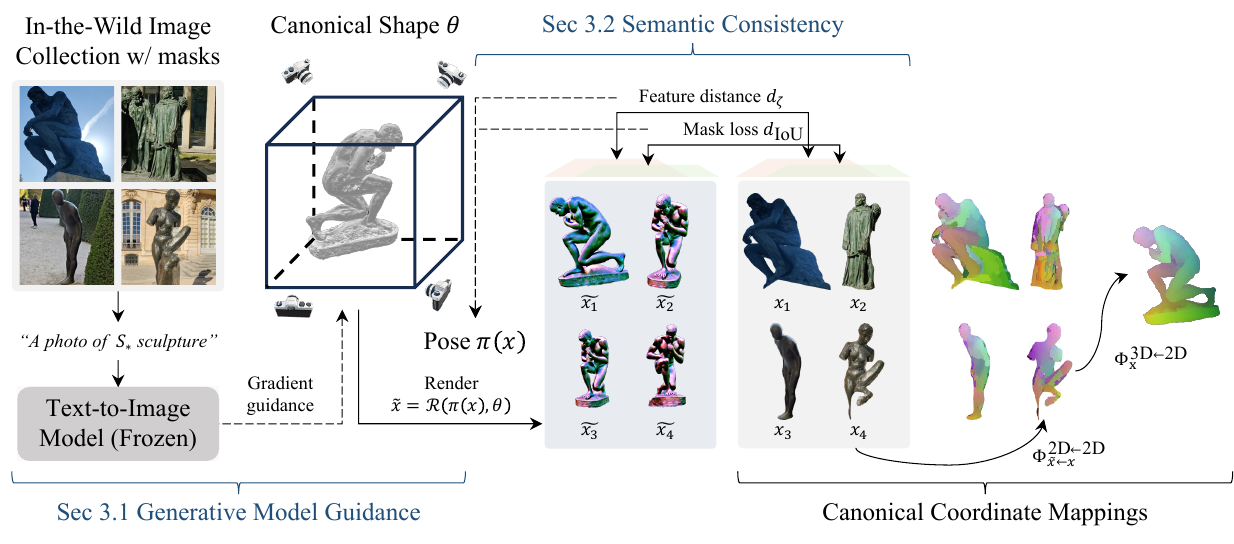}
    \vspace{-1.5em}
    \caption{\textbf{Pipeline.} Given a collection of in-the-wild images capturing similar objects as inputs, we develop a framework that ``congeals'' these images in 3D. The core representation consists of a canonical 3D shape that captures the geometric structure shared among the inputs, together with a set of coordinate mappings that register the input images to the canonical shape. 
    The framework utilizes the prior knowledge of plausible 3D shapes from a generative model, and aligns images in the semantic space using pre-trained semantic feature extractors.
    } 
    \label{fig:method}
    \vspace{-1em}
\end{figure}

\noindent \textbf{Problem Formulation.}
We formulate the problem of \task as follows. Given a set of $N$ object-centric images $\mathcal{D} = \{x_n\}_{n=1}^N$ that captures objects sharing semantic components, \eg, objects from one category, we seek to align the object instances in these images into a canonical 3D representation, \eg, NeRF~\cite{mildenhall2021nerf}, parameterized by $\theta$. We refer to the coordinate frame of this 3D representation as the canonical frame. 
We also recover the camera pose of each observation $x\in\mathcal{D}$ in the canonical frame, denoted using a pose function $\pi: x \mapsto (\xi, \kappa)$ where $\xi$ represents the object pose in $\mathrm{SE}(3)$ and $\kappa$ is the camera intrinsic parameters. 
We assume access to instance masks, which can be easily obtained using an off-the-shelf segmentation method~\cite{kirillov2023segment}. 

The 3D representation should be consistent with the physical prior of objects in the natural world, and with input observations both geometrically and semantically. These constraints can be translated into an optimization problem:
\begin{equation}
    \max_{\pi, \theta} p_\Theta(\theta),
    \text{s.t.}\; x = \mathcal{R}(\pi(x), \theta), \forall x\in \mathcal{D},
    \label{eq:problem}
\end{equation}
where $p_\Theta$ is a prior distribution for the 3D representation parameter $\theta$ that encourages physically plausible solutions, $\mathcal{R}$ is a predefined rendering function that enforces geometric consistency, and the equality constraint on image reconstruction enforces compliance with input observations. 

In the following sections, we will describe an instantiation of the 3D prior $p_\Theta$ (\cref{subsect:generative_guidance}), an image distance function that helps enforce the equality constraint (\cref{subsect:discriminative_guidance}), followed by the \task optimization (\cref{subsect:optimization}). 

\subsection{3D Guidance from Generative Models}
\label{subsect:generative_guidance}
As illustrated in the left part of \Cref{fig:method}, we extract the prior knowledge for 3D representations $p_\Theta(\cdot)$ from a pre-trained text-to-image (T2I) model such as Stable-Diffusion~\cite{rombach2022high}. 
DreamFusion~\cite{poole2022dreamfusion} proposes to turn a text prompt $y$ into a 3D representation $\theta$ using the following Score Distillation Sampling (SDS) objective, leveraging a T2I diffusion model with frozen parameters $\phi$,
\begin{equation}
    \min_\theta 
    \mathbb{E}_{x\in\mathcal{D}(\theta)}\mathcal{L}_\text{diff}^\phi(x, y). 
    \label{eq:dreamfusion}
\end{equation}
Here $\mathcal{D}(\theta) :=\{\mathcal{R}(\pi, \theta) \mid \pi\sim p_\Pi(\cdot)\}$ contains images rendered from the 3D representation $\theta$ under a prior camera distribution $p_\Pi(\cdot)$, and $\mathcal{L}_\text{diff}^\phi$ is the training objective of image diffusion models specified as follows:
\begin{equation}
    \mathcal{L}_\text{diff}^\phi(x, y)
    :=\mathbb{E}_{t\sim\mathcal{U}(\left[0, 1\right]),\epsilon\sim\mathcal{N}(\mathbf{0}, I)} \left[\omega(t)\|
    \epsilon_\phi(
    \alpha_t x + \sigma_t \epsilon, y, t) - \epsilon\|_2^2\right],
    \label{eqn:loss_diff}
\end{equation}
where $\epsilon_\phi$ is the pre-trained denoising network, $\omega(\cdot)$ is the timestep-dependent weighting function, $t$ is the diffusion timestep and and $\alpha_t, \sigma_t$ are timestep-dependent coefficients from the diffusion model schedule. 

The above loss can be used to guide the optimization of a 3D representation $\theta$, whose gradient is approximated by 
\begin{equation}
\nabla_\theta \mathcal{L}_\text{diff}^\phi(x=\mathcal{R}(\xi, \kappa, \theta), y)
\approx \mathbb{E}_{t, \epsilon}\left[ \omega(t)(
    \epsilon_\phi(
    \alpha_t x + \sigma_t \epsilon, y, t) - \epsilon)\frac{\partial x}{\partial \theta}\right],
    \label{eqn:sds} 
\end{equation}
where $\xi$ and $\kappa$ are the extrinsic and intrinsic camera parameters, respectively.
The derived gradient approximation is adopted by later works such as MVDream~\cite{shi2023mvdream}, which we use as the backbone. 

The original SDS objective is optimizing for a text-conditioned 3D shape with a user-specified text prompt $y$ and does not consider image inputs. 
Here, we use the technique from Textual Inversion~\cite{gal2022image} to recover the most suitable text prompt $y^*$ that explains input images, defined as follows:
\begin{equation}
    y^* = \arg\min_y \mathbb{E}_{x\in\mathcal{D}}\mathcal{L}_\text{diff}^\phi(x, y).
    \label{eq:textual_inversion}
\end{equation}

\Cref{eq:dreamfusion} and \Cref{eq:textual_inversion} differ in that both the sources of the observations $x$ (an infinite dataset of rendered images $\mathcal{D}(\theta)$ for the former, and real data $\mathcal{D}$ for the latter) and the parameters being optimized over ($\theta$ and $y$, respectively).
In our framework, we incorporate the real image information to the SDS guidance via first solving for $y^*$ (\Cref{eq:textual_inversion}) and keep it frozen when optimizing for $\theta$ (\Cref{eq:dreamfusion}). The diffusion model parameter $\phi$ is frozen throughout the process, requiring significantly less memory compared to the alternative of integrating input image information via finetuning $\phi$ as in DreamBooth3D~\cite{raj2023dreambooth3d}.

\subsection{Semantic Consistency from Deep Features}
\label{subsect:discriminative_guidance}
The generative model prior from \cref{subsect:generative_guidance} effectively constrains the search space for the solutions. However, the objectives from \Cref{eq:dreamfusion,eq:textual_inversion} use the input image information only indirectly, via a text embedding $y^*$.
To explain the relative geometric relation among input images, we explicitly recover the pose of each input image \wrt $\theta$, as illustrated in \Cref{fig:method} (middle) and as explained below.

To align input images, we use an image distance metric defined by semantic feature dissimilarity.
In particular, pre-trained deep models such as DINO~\cite{caron2021emerging,oquab2023dinov2}
have been shown to be effective semantic feature extractors. Denote such a model as $f$ parameterized by $\zeta$. The similarity of two pixel locations $u_1$ and $u_2$ from two images $x_1$ and $x_2$, respectively, can be measured with
\begin{equation}
     d_\zeta^{u_1,u_2}(x_1, x_2)
 := 1 - \frac{\langle \left[f_\zeta(x_1)\right]_{u_1}, \left[f_\zeta(x_2)\right]_{u_2} \rangle}{\|\left[f_\zeta(x_1)\right]_{u_1}\|_2 \|\left[f_\zeta(x_2)\right]_{u_2}\|_2},
 \label{eq:distance_pixelwise}
\end{equation}
where $\left[\cdot\right]$ is an indexing operator.
It thereafter defines an image distance function
\begin{equation}
 \|x_1 - x_2\|_{d_\zeta}
 := \frac{1}{HW}\sum_u d_\zeta^{u, u}(x_1, x_2),
    \label{eq:distance}
\end{equation}
where $x_1$ and $x_2$ have resolution $H\times W$, and the sum is over all image coordinates. 

The choice of semantic-aware image distance, instead of photometric differences as in the classical problem setting of multiview 3D reconstruction~\cite{schonberger2016structure,wang2021neus,yariv2021volume}, leads to solutions that maximally align input images to the 3D representation with more tolerance towards variance in object shape, texture, and environmental illuminations among input images, which is crucial in our problem setting.

\subsection{Optimization}
\label{subsect:optimization}
\paragraph{The Canonical Shape and Image Poses.}
Combining \cref{subsect:generative_guidance,subsect:discriminative_guidance}, we convert the original problem in \Cref{eq:problem} into
\begin{equation}
\begin{aligned}
    \min_{\pi,\theta} \underbrace{\mathbb{E}_{x\in{\mathcal{D}}(\theta)}\mathcal{L}_\text{diff}^\phi(x, y^*)}_\text{generative model guidance} + \lambda\underbrace{\mathbb{E}_{x\in\mathcal{D}} \|\mathcal{R}(\pi(x), \theta) - x\|_d}_\text{data reconstruction},
\end{aligned}
\label{eq:problem_joint}
\end{equation}
where $y^*$ come from \Cref{eq:textual_inversion} and $\lambda$ is a loss weight.
Compared to \Cref{eq:textual_inversion}, here the first term instantiates the generative modeling prior and the second term is a soft constraint of reconstructing input observations. Specifically, $d = \lambda_\zeta d_\zeta + \lambda_\text{IoU} d_\text{IoU}$, where $d_\zeta$ is the semantic-space distance metric from \cref{subsect:discriminative_guidance}, and $d_\text{IoU}$ is the Intersection-over-Union (IoU) loss for masks, $\| m_1 - m_2\|_{d_\text{IoU}} := 1 - (\|m_1 \odot m_2\|_1)/(\|m_1\|_1 + \|m_2\|_1 - \|m_1 \odot m_2\|_1)$,
where $m_1$ and $m_2$ are image masks, which in \Cref{eq:problem_joint} are set to be the mask rendering and the instance mask for $x$. 
The use of both $d_\zeta$ and $d_\text{IoU}$ tolerates shape variance among input instances. 

For the shape representation, 
we follow NeRF~\cite{mildenhall2021nerf} and use neural networks $\sigma_\theta: \mathbb{R}^3\rightarrow \mathbb{R}$ and $c_\theta: \mathbb{R}^3\rightarrow \mathbb{R}^3$ to map a 3D spatial coordinate to a density and an RGB value, respectively. The rendering operation $\mathcal{R}$ is the volumetric rendering operation specified as follows:
\begin{equation}
    \mathcal{R}(r, \xi, \theta; c_\theta)= \int T(t)\sigma_\theta(\xi r(t)) c_\theta(\xi r(t)) \;\mathrm{d}t,
    \label{eq:nerf}
\end{equation}
where $T(t) = \exp\left(-\int \sigma_\theta(r(t^\prime))\mathrm{d}t^\prime\right)$, $r: \mathbb{R}\rightarrow\mathbb{R}^3$ is a ray shooting from the camera center to the image plane, parameterized by the camera location and the ray's direction, and $\xi$ is the relative pose that transforms the ray from the camera frame to the canonical frame.

\paragraph{Forward Canonical Coordinate Mappings.}
After the above optimization,
each image $x$ from the input image collection can be ``congealed'' to the shape $\theta$ via a \emph{canonical coordinate mapping}, \ie, a forward warping operation $\Phi_x^\text{fwd}: \mathbb{R}^2 \rightarrow \mathbb{R}^3$ that maps a 2D image coordinate to a 3D coordinate in the canonical frame of reference as illustrated in \Cref{fig:method}.
$\Phi_x^\text{fwd}$ consists of the following two operations. 

First, we warp a coordinate $u$ from the real image $x$ to the rendering of the canonical shape under its pose $\pi(x)$, denoted as $\tilde{x} := \mathcal{R}(\pi(x), \theta)$. Specifically, 
\begin{equation}
    \Phi_{\tilde{x} \leftarrow x}^{\text{2D}\leftarrow\text{2D}}(u)
    := \arg\min_{\tilde{u}}
    d_\zeta^{\tilde{u},u}(\tilde{x}, x) + \lambda_{\ell_2} \|\tilde{u} - u\|_2^2 + \lambda_\text{smooth}\mathcal{L}_\text{smooth}(\tilde{u}, u),
\label{eq:forward_2d_to_2d}
\end{equation}
where $d_\zeta$ follows \Cref{eq:distance_pixelwise}, the 2D coordinates $u$ and $\tilde{u}$ are normalized into range $\left[0, 1\right]$ before computing the $\ell_2$ norm, the smoothness term $\mathcal{L}_\text{smooth}$ is specified in 
\cref{app:regularization}, and $\lambda_{\ell_2}$ and $\lambda_{smooth}$ are scalar weights.
This objective searches for a new image coordinate $\tilde{u}$ (from the rendering $\tilde{x}$) that shares a semantic feature similar to $u$ (from the real image $x$), and ensures that $\tilde{u}$ stays in the local neighborhood of $u$ via a soft constraint of the coordinate distance. 
Afterward, a 2D-to-3D operation takes in the warped coordinate from above and outputs its 3D location in the normalized object coordinate space (NOCS)~\cite{wang2019normalized} of $\theta$: %
\begin{equation}
  \Phi_x^{\text{3D}\leftarrow\text{2D}}(\tilde{u}): = \left[\mathcal{R}_\text{NOCS}(\pi(x), \theta)\right]_{\tilde{u}},
    \label{eq:forward_2d_to_3d}
\end{equation}
where $\mathcal{R}_\text{NOCS}$ is identical to $\mathcal{R}$ from \Cref{eq:nerf}, but replacing the color field $c_\theta$ with a canonical object coordinate field, $c_\text{NOCS}: \mathbb{R}^3\rightarrow \mathbb{R}^3, p \mapsto (p - p_\text{min}) / (p_\text{max} - p_\text{min})$,
where $p_\text{min}$ and $p_\text{max}$ are the two opposite corners of the canonical shape's bounding box. These bounding boxes are determined by the mesh extracted from the density neural field $\sigma_\theta$ using the Marching Cube~\cite{lorensen1998marching} algorithm.

Combining the above, given an input image coordinate $u$, $\Phi_x^\text{fwd}(u) := \Phi_x^{\text{3D}\leftarrow\text{2D}} \circ  \Phi_{\tilde{x}\leftarrow x}^{\text{2D}\leftarrow\text{2D}} (u)$ identifies a 3D location in the canonical frame corresponding to $u$.  

\paragraph{Reverse Canonical Coordinate Mappings.}
Each image can be ``uncongealed'' from the canonical shape using $\Phi_x^\text{rev}: \mathbb{R}^3 \rightarrow \mathbb{R}^2$, which is the reverse operation of $\Phi_x^\text{fwd}(u)$ and is approximately computed via nearest-neighbor inversion as explained below.

Given a 3D location within a unit cube, $p\in \left[0, 1\right]^3$,
$\Phi_x^\text{rev}(p) := \Phi_{x\leftarrow \tilde{x}}^{\text{2D}\leftarrow\text{2D}} \circ\Phi_{x}^{\text{2D}\leftarrow\text{3D}}(p)$. In particular, 
\begin{equation}
    \Phi_x^{\text{2D}\leftarrow\text{3D}}(p) := \arg\min_{\tilde{u}} \|p - \Phi_x^{\text{3D}\leftarrow\text{2D}}(\tilde{u})\|_2
    \label{eq:reverse_3d_to_2d}
\end{equation}

is an operation that takes in a 3D coordinate $p$ in the canonical frame and searches for a 2D image coordinate whose NOCS value is the closest to $p$, and $\Phi_{x\leftarrow \tilde{x}}^{\text{2D}\leftarrow\text{2D}}$ is computed via inverting $\Phi_{\tilde{x}\leftarrow x}^{\text{2D}\leftarrow\text{2D}}$ from \Cref{eq:forward_2d_to_2d},
\begin{equation}
\Phi_{x\leftarrow \tilde{x}}^{\text{2D}\leftarrow\text{2D}}(\tilde{u}) := \arg\min_u \|\tilde{u} - \Phi_{\tilde{x}\leftarrow x}^{\text{2D}\leftarrow\text{2D}}(u)\|_2.
    \label{eq:reverse_2d_to_2d} 
\end{equation}

\begin{figure}[t]
    \centering
    \includegraphics[width=\linewidth]{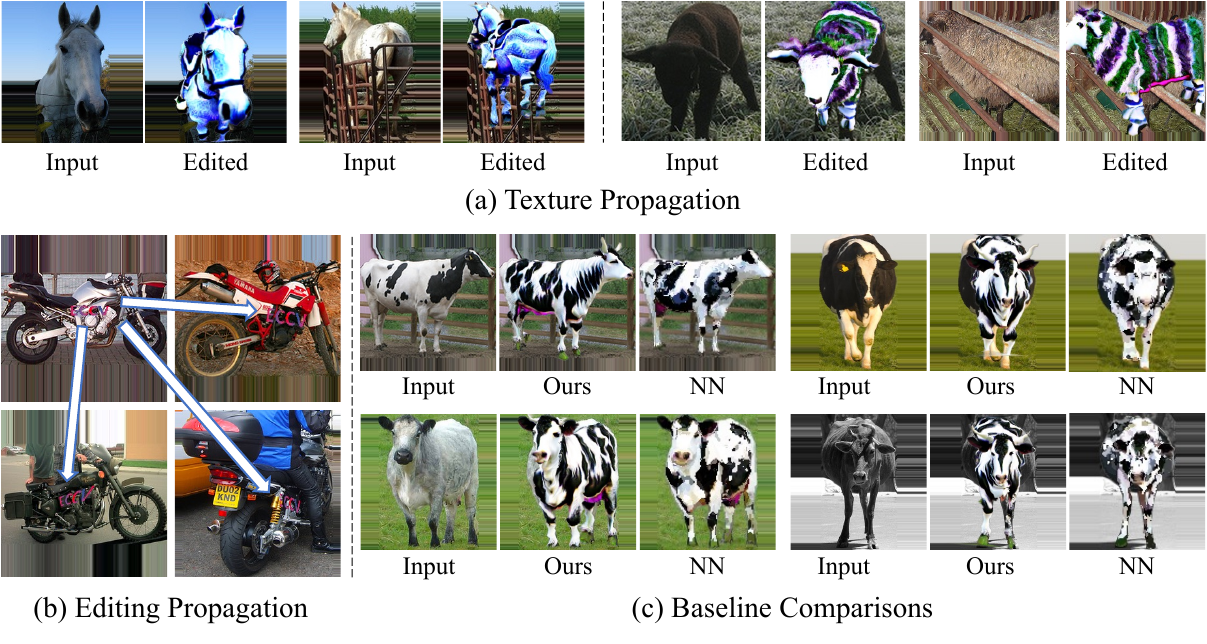}
    \vspace{-2em}
    \caption{\textbf{Image Editing.}
    Our method propagates (a) texture and (b) regional editing to real images, (c) achieving smoother results compared to the nearest-neighbor (NN) baseline thanks to the 3D geometric reasoning. 
    } 
\label{fig:rebuttal_editing}
\end{figure}

In summary, the above procedure establishesthe 2D-3D correspondence between an input image $x$ and the canonical shape via $\Phi_x^\text{fwd}$, and defines the dense 2D-2D correspondences between two images $x_1, x_2$ via $\Phi_{x_2}^\text{rev}\circ \Phi_{x_1}^\text{fwd}$ which enables image editing (\Cref{fig:rebuttal_editing}). 
The full framework is described in \cref{alg}.

\subsection{Implementation Details}

\begin{wrapfigure}{r}{0.5\textwidth}
\begin{minipage}{\linewidth}
\begin{algorithmic}[1]
\vspace{-2.5em}
\scriptsize
\Procedure{RUN}{$\mathcal{D}=\{x_n\}_{n=1}^N$}
    \State $y^* \gets$ Solution to \Cref{eq:textual_inversion}
    \State Optimize $\theta$ with \Cref{eq:problem_joint}
    \State Sample pose candidates $\{\xi_i\}_i$
    \For{$n \gets 1$ to $N$} \Comment{Pose initialization}
        \State $\pi(x_n)\gets \arg\min_{\xi_i} \|\mathcal{R}(\xi, \theta) - x_n\|_{d_\zeta}$
    \EndFor
    \State Optimize $\pi(x_n)$ with \Cref{eq:problem_joint} for all $n$
    \State Determine $\Phi_{x_n}^\text{fwd}$ and $\Phi_{x_n}^\text{rev}$ for all $n$
    \State \textbf{return} $\theta$, $\pi$, $\{\Phi_{x_n}^\text{fwd}\}_{n=1}^N$,$\{\Phi_{x_n}^\text{rev}\}_{n=1}^N$
\EndProcedure
\end{algorithmic}
\captionof{algorithm}{\textbf{Overview.}}
\vspace{-1em}
\label{alg}
\end{minipage}
\end{wrapfigure}
Input images are cropped with the tightest bounding box around the foreground masks. The masks come from dataset annotations, if available, or from Grounded-SAM~\cite{ren2024grounded,kirillov2023segment}, an off-the-shelf segmentation model, for all Internet images.

Across all experiments, we optimize for $y^*$ (\cref{alg}, line 2) for $1,000$ iterations using an AdamW~\cite{loshchilov2018decoupled} optimizer with learning rate $0.02$ and weight decay $0.01$. We optimize for $\theta$ (line 3) with $\lambda = 0$ for $10,000$ iterations, with AdamW and learning rate $0.001$.
The NeRF model $\theta$ has 12.6M parameters. It is frozen afterwards and defines the coordinate frame for poses. 

Since directly optimizing poses and camera parameters with gradient descents easily falls into local minima~\cite{lin2021barf}, we initialize $\pi$ using an analysis-by-synthesis approach (\cref{alg}, line 5-7). Specifically, we parameterize the camera intrinsics using a pinhole camera model with a scalar Field-of-View (FoV) value, and sample the camera parameter $(\xi, \kappa)$ from a set of candidates determined by an exhaustive combination of 3 FoV values, 16 azimuth values, and 16 elevation values uniformly sampled from $\left[15\degree, 60\degree\right]$, $\left[-180\degree, 180\degree\right]$, and $\left[-90\degree, 90\degree\right]$, respectively. 
In this pose initialization stage, all renderings use a fixed camera radius and are cropped with the tightest bounding boxes, computed using the rendered masks, before being compared with the real image inputs. 
We then select the candidate with the lowest error measured by the image distance function $d$ from \cref{subsect:discriminative_guidance}, with $\lambda_\zeta = 1$ and $\lambda_\text{IoU} = 0$. 

After pose initialization, we use the $\mathfrak{se}(3)$ Lie algebra for camera extrinsics parameterization following BARF~\cite{lin2021barf}, and optimize for the extrinsics and intrinsics of each input image (\cref{alg}, line 8), with $\lambda_\zeta = 0$ and $\lambda_\text{IoU} = 1$, for $1,000$ iterations with the Adam~\cite{kingma2015adam} optimizer and learning rate $0.001$. Since $\theta$ is frozen, the optimization effectively only considers the second term from \Cref{eq:problem_joint}. 
Finally, to optimize for the canonical coordinate mappings (\cref{alg}, line 9), for each input image, we run $4,000$ iterations for \Cref{eq:forward_2d_to_2d} with AdamW and learning rate $0.01$.
All experiments are run on a single 24GB A5000 GPU.

\begin{figure}[t]
    \centering
    \includegraphics[width=\linewidth]{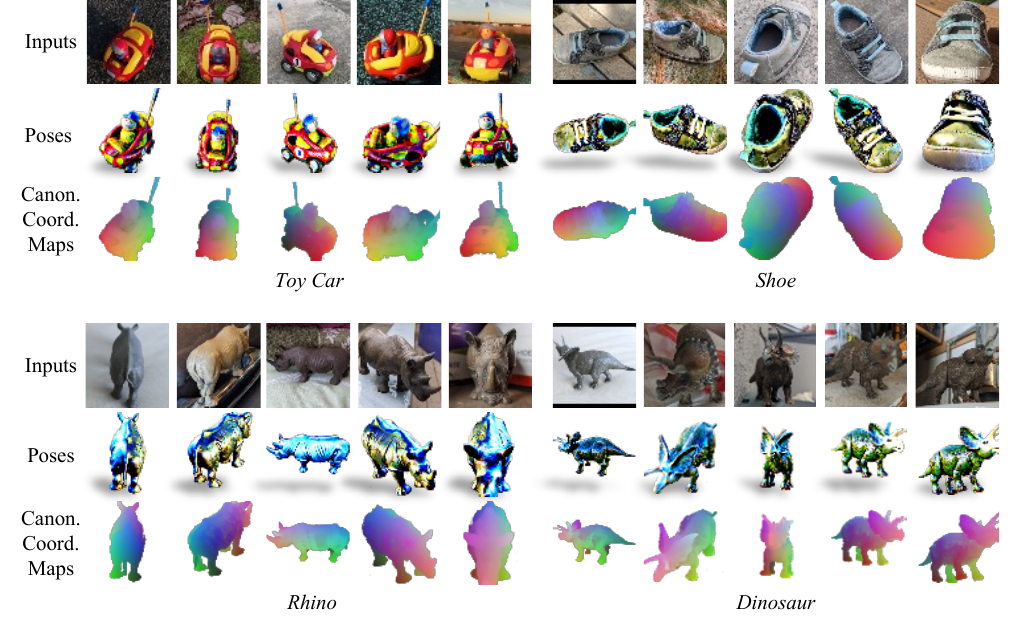}
    \vspace{-2em}
    \caption{\textbf{Pose Estimation from Multi-Illumination Captures.} The figure shows 4 example scenes from the NAVI dataset, displaying the real image inputs, canonical shapes under estimated poses, and the canonical coordinate maps.}
    \vspace{-1em}
    \label{fig:exp_navi}
\end{figure}

\begin{table}[t]
\centering
\scriptsize
\setlength{\tabcolsep}{5pt}
\begin{tabular}{llcccc}
\toprule
\multirow{2}{*}{Labels} & \multirow{2}{*}{Methods} & \multicolumn{2}{c}{Rotation\textdegree $\downarrow$} 
 & \multicolumn{2}{c}{Translation$\downarrow$}
 \\ 
 \cmidrule(r){3-4}\cmidrule(r){5-6}
 & & \multicolumn{1}{c}{$S_C$} & \multicolumn{1}{c}{$\sim S_C$} &\multicolumn{1}{c}{$S_C$} 
 & \multicolumn{1}{c}{$\sim S_C$} 
 \\ \midrule
\multirow{3}{*}{Pose} & NeROIC~\cite{kuang2022neroic} & $42.11$ & - & $\boldsymbol{0.09}$ & - \\
& NeRS~\cite{zhang2021ners} & $122.41$ & $123.63$ & $0.49$ & $0.52$ \\
& SAMURAI~\cite{boss2022samurai} & $\boldsymbol{26.16}$ & $\boldsymbol{36.59}$ & $0.24$ & $\boldsymbol{0.35}$ \\\midrule
\multirow{5}{*}{None} & GNeRF~\cite{meng2021gnerf} & $93.15$ & $80.22$ & $1.02$ & $1.04$ \\
& PoseDiffusion~\cite{wang2023posediffusion}&$46.79$ & $46.34$ &$0.81$ & $0.90$ \\
& Ours (3 seeds) & $\boldsymbol{26.97}$\xpm{2.24} & $\boldsymbol{32.56}$\xpm{2.90} & $\boldsymbol{0.40}$\xpm{0.01} & $\boldsymbol{0.41}$\xpm{0.04} \\
 \cmidrule(r){2-6}
& Ours (No Pose Init) & $53.45$ &$57.87$ & $0.97$ & $0.96$ \\
& Ours (No IoU Loss) &$31.29$ & $31.15$ & $0.87$& $0.85$ \\
\bottomrule
\end{tabular}
\vspace{0.5em}
\caption{\textbf{Pose Estimation from Multi-Illumination Image Captures.} 
Our method performs better than both GNeRF and PoseDiffusion with the same input information, and on par with SAMURAI which additionally assumes camera pose direction as inputs. 
Different random seeds lead to different canonical shapes, but our method is robust to such variations. $\pm$ denotes means followed by standard deviations.
}
\vspace{-1em}
\label{tab:pose}
\end{table}

\begin{figure}[t]
    \centering
    \includegraphics[width=\linewidth]{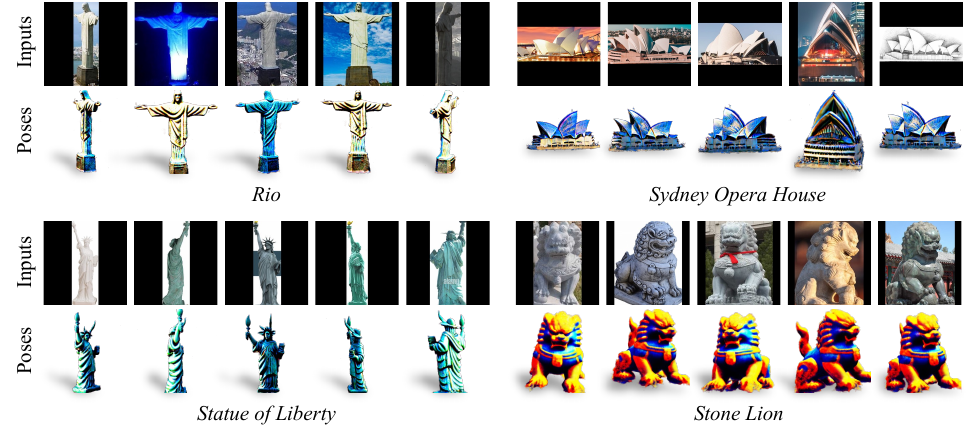}
    \vspace{-2em}
    \caption{\textbf{Pose Estimation for Tourist Landmarks.} 
    This is a challenging problem setting due to the varying viewpoints and lighting conditions, and the proposed method can successfully align online tourist photos taken at different times and possibly at different geographical locations, into one canonical representation.}
    \label{fig:exp_localization}
    \vspace{-1em}
\end{figure}

\begin{figure}[t]
    \centering
    \includegraphics[width=\linewidth]{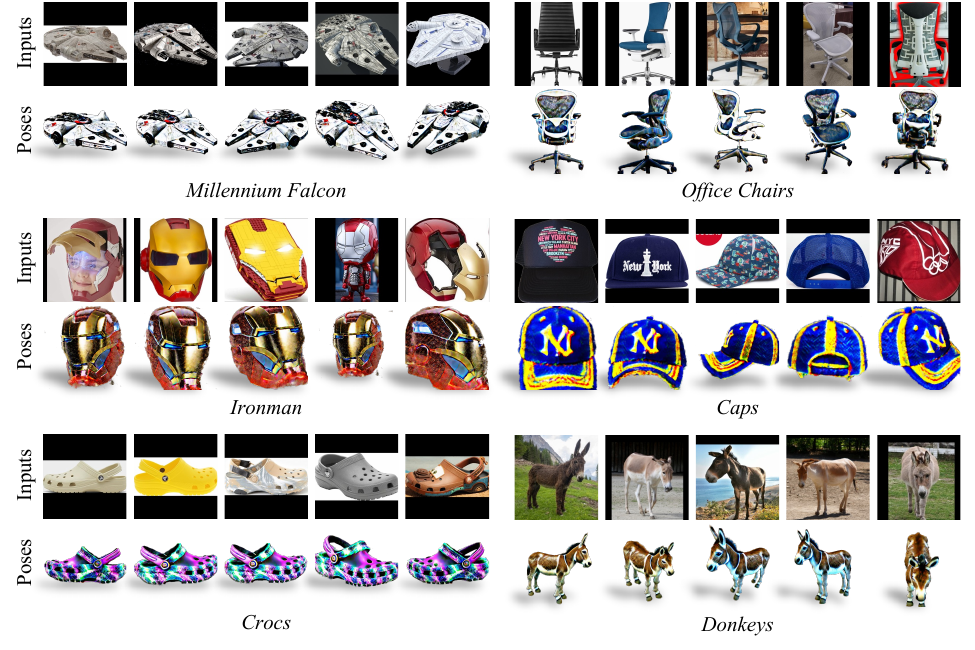}
    \vspace{-2em}
    \caption{\textbf{Object Alignment from Internet Images.} 
    Results of an online image search may contain various appearances, identities, and articulated poses of the object. Our method can successfully associate these in-the-wild images with one shared 3D space.
    }
    \vspace{-1em}
    \label{fig:exp_retrieval}
\end{figure}
\begin{figure}[t]
    \centering
    \includegraphics[width=\linewidth]{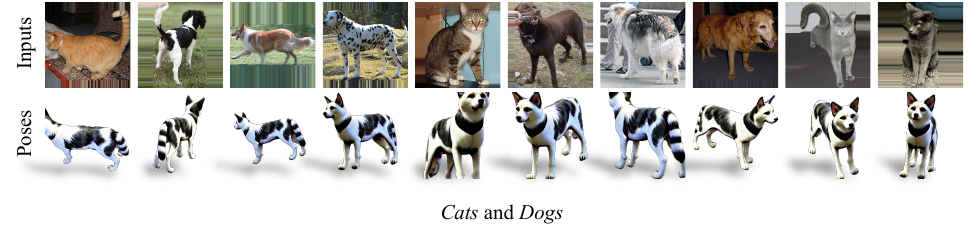}
    \vspace{-2em}
    \caption{\textbf{Cross-Category Results.} 
    The method can associate images from different categories, such as cats and dogs, by leveraging a learned average shape.
    }
\label{fig:exp_cross_category}
\end{figure}

\begin{figure}[t]
    \centering
    \includegraphics[width=\linewidth]{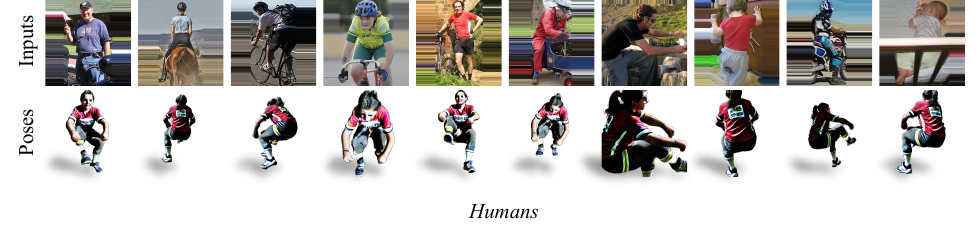}
    \vspace{-2em}
    \caption{\textbf{Results on Deformable Objects.} 
    The method can be applied to images with highly diverse articulated poses and shapes as shown in the examples above.
    } 
    \vspace{-1em}
\label{fig:rebuttal_human}
\end{figure}

\section{Experiments}
\label{sect:experiments}

In this section, we first benchmark the pose estimation performance of our method on in-the-wild image captures (\cref{subsect:exp_pose}), and then show qualitative results on diverse input data and demonstrate applications such as image editing (\cref{subsect:applications}).

\subsection{Pose Estimation}
\label{subsect:exp_pose}

\paragraph{Dataset.}
While our method does not require input object instances to be identical, we use a dataset of multi-illumination object-centric image captures with ground truth camera poses for evaluation. Specifically, we use the in-the-wild split of the NAVI~\cite{jampani2023navi} dataset, which contains 35 object image collections in its official release. 
Each image collection contains an average of around 60 casual image captures of an object instance placed under different illumination conditions, backgrounds and cameras with ground truth poses. 

We use identical hyperparameters for all scenes. We do not introduce additional semantic knowledge for objects contained in the scene and use a generic text prompt, ``a photo of sks object'', for initialization for all scenes. The text embeddings corresponding to the tokens for ``sks object'' are being optimized using \Cref{eq:textual_inversion} with the embeddings for others being frozen. 
For each scene, it takes around 1 hr to optimize for the NeRF model, 15 min for pose initialization, and 45 min for pose optimization.

\paragraph{Baselines.}

We compare with several multiview reconstruction baselines. In particular, NeROIC~\cite{kuang2022neroic} uses the poses from COLMAP, and NeRS~\cite{zhang2021ners} and SAMURAI~\cite{boss2022samurai} require initial camera directions. 
GNeRF~\cite{meng2021gnerf} is a pose-free multiview 3D reconstruction method that is originally designed for single-illumination scenes, and is adapted as a baseline using the same input assumption as ours.
PoseDiffusion~\cite{wang2023posediffusion} is a learning-based framework that predicts relative object poses, using ground truth pose annotations as training supervision. 
The original paper takes a model pre-trained on CO3D~\cite{reizenstein2021common} and evaluates the pose prediction performance in the wild, and we use the same checkpoint for evaluation. 

\paragraph{Metrics.}
The varying illuminations pose challenges to classical pose estimation methods such as COLMAP~\cite{schonberger2016structure}. We use the official split of the data which partitions the 35 scenes into 19 scenes where COLMAP converges ($S_C$ in \Cref{tab:pose}), and 16 scenes where COLMAP fails to converge ($\sim S_C$). 
Following~\cite{jampani2023navi}, we report the absolute rotation and translation errors using Procrustes analysis~\cite{gower2004procrustes}, where for each scene, the predicted camera poses are aligned with the ground truth pose annotations using a global transformation before computing the pose metrics. 

\paragraph{Results.}
Handling different illumination conditions is challenging for all baselines using photometric-reconstruction-based optimization~\cite{zhang2021ners,kuang2022neroic,boss2022samurai} even with additional information for pose initialization. As shown in \Cref{tab:pose}, our approach significantly outperforms both GNeRF and PoseDiffusion and works on par with SAMURAI which requires additional pose initialization. We run our full pipeline with 3 random seeds and observe a consistent performance across seeds.
As shown in \Cref{fig:exp_navi} results, across a wide range of objects captured in this dataset, our method accurately estimates the poses for the input images and associates all inputs together in 3D via the canonical coordinate maps. 

\begin{figure}[t]
    \centering
    \includegraphics[width=\linewidth]{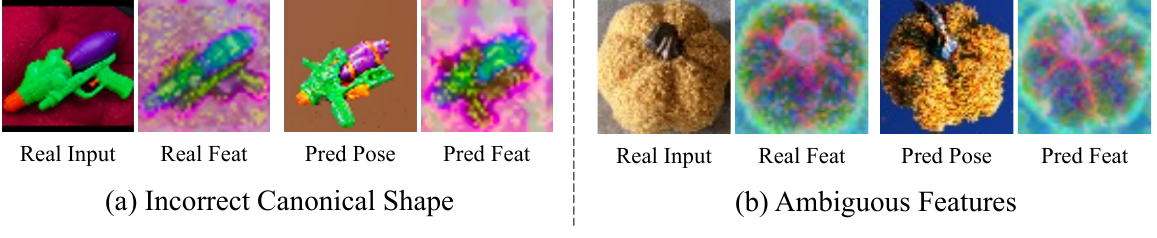}
    \vspace{-2em}
    \caption{\textbf{Failure Modes.} 
    Our method inherits the failure from (a) canonical shape optimization and (b) pre-trained feature extractors.
    } 
    \vspace{-1em}
\label{fig:rebuttal_pose_failure}
\end{figure}

\paragraph{Ablations.}
\Cref{tab:pose} also shows ablation for the pose fitting objectives. The initialization is critical (``No Pose Init''), which is expected as pose optimization is susceptible to local optima~\cite{lin2021barf}.
``No IoU Loss'', which is equivalent to using the initialized poses as final predictions, also negatively affects the performance.

\subsection{Applications}
\label{subsect:applications}

We show qualitative results on various in-the-wild image data. Inputs for \Cref{fig:exp_localization,fig:exp_retrieval} are crawled with standard online image search engines and are CC-licensed, each consisting of $50$ to $100$ images. Inputs for \Cref{fig:exp_cross_category,fig:rebuttal_human} come from the SPair-71k dataset~\cite{min2019spair}. We use identical hyperparameters for all datasets, except for text prompt initialization where we use a generic description of the object, \eg, ``a photo of sks sculpture'', or ``a photo of cats plus dogs'' for \Cref{fig:exp_cross_category}.

\paragraph{Single-Instance.}
\Cref{fig:exp_localization} shows the result on Internet photos of tourist landmarks, which may contain a large diversity in illuminations (\eg, the Rio) and styles (\eg, the sketch of the Sydney Opera House).
The proposed method can handle such variances and align these photos or art pieces, which are abundant from the Internet image database, to the same canonical 3D space and recover the relative camera poses.

\paragraph{Cross-Instance, Single-Category.}
Internet images from generic objects may contain more shape and texture variations compared to the landmarks. \Cref{fig:exp_retrieval} shows results for various objects. 
Our framework consistently infers a canonical shape from the input images to capture the shared semantic components being observed. For example, the corresponding semantic parts of faces of humans and the Ironman are identified as similar and are aligned with each other. 

\paragraph{Cross-Category.}
The method does not make an assumption on the category of inputs. Given cross-category inputs, such as a mixture of cats and dogs as shown in \Cref{fig:exp_cross_category}, the method effectively infers an average shape as an anchor to further reason about the relative relation among images from different categories.

\paragraph{Inputs with Deformable Shapes.}
To test the robustness of the method, we run the pipeline on images of humans with highly diverse poses. \Cref{fig:teaser,fig:rebuttal_human} show that the method assigns plausible poses to the inputs despite the large diversity of shapes and articulated poses contained in the inputs.

\paragraph{Image Editing.}
The proposed method finds image correspondence and can be applied to image editing, as shown in \Cref{fig:rebuttal_editing} (a-b).
\Cref{fig:rebuttal_editing} (c) shows that our method obtains more visually plausible results compared to the Nearest-Neighbor (NN) baseline using the same DINO features. The baseline matches features in 2D for each pixel individually and produce noisy results, as discussed in \cref{app:analysis}. Quantitative evaluation of correspondence matching is included in \cref{app:correspondence}.

\subsection{Failure Modes}
\label{subsect:analysis}

We have identified two failure modes of the proposed method: (1) incorrect shapes from the generative model distillation process, \eg, the incorrect placement of the water gun handle from \Cref{fig:rebuttal_pose_failure} (a), and (2) incorrect poses due to feature ambiguity, \eg, the pumpkin is symmetric and DINO features cannot disambiguate sides from \Cref{fig:rebuttal_pose_failure} (b). 

\section{Conclusion}

We have introduced \task, 3D-aware alignment for 2D images capturing semantically similar objects. Our proposed framework leverages a canonical 3D representation that encapsulates geometric and semantic information and, through optimization, fuses prior knowledge from a pre-trained image generative model and semantic information from input images. We show that our model achieves strong results on real-world image datasets under challenging identity, illumination, and background conditions.
\paragraph{Acknowledgments.} We thank Chen Geng and Sharon Lee for their help in reviewing the manuscript. This work is in part supported by NSF RI \#2211258, \#2338203, and ONR MURI N00014-22-1-2740.

\bibliographystyle{splncs04}
\bibliography{main}
\appendix

\begin{figure}[t]
    \centering
    \includegraphics[width=0.97\linewidth]{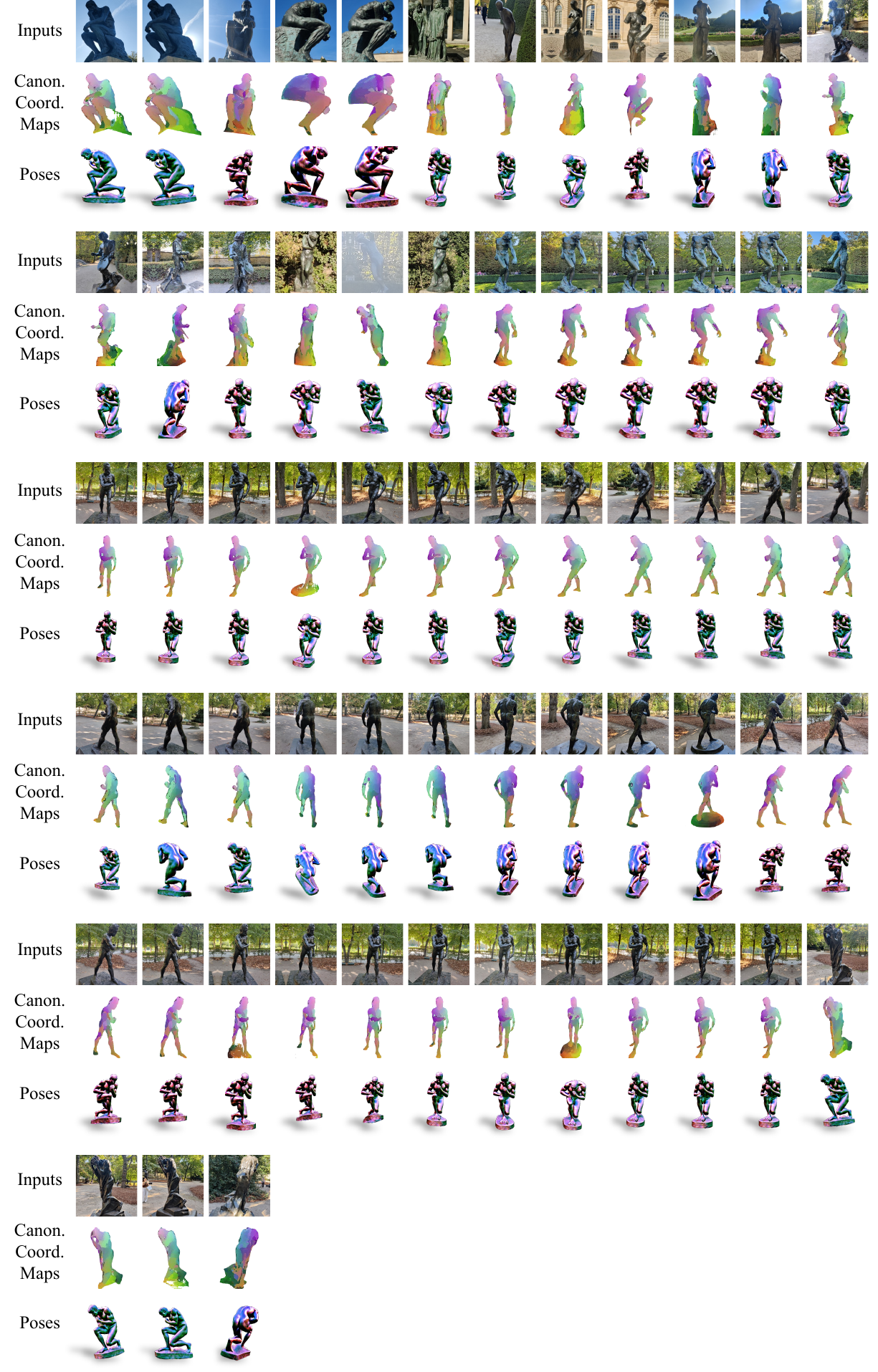}
    \caption{\textbf{Results on the Sculpture Dataset.}}
    \label{fig:rodin_full}
\end{figure}

\section{Additional Qualitative Results}
\label{app:rodin_full}
The complete set of input images used for the sculpture dataset from \Cref{fig:teaser} and the corresponding results are shown in \Cref{fig:rodin_full}. Images come from a personal photo collection. 

\section{Implementation Details}
\label{app:regularization}
\paragraph{Feature Extractors.}

We use the ViT-G/14 variant of DINO-V2~\cite{oquab2023dinov2} as the feature extractor $f_\zeta$ from \cref{subsect:discriminative_guidance} and extract tokens from its final layer for all quantitative experiments. For qualitative results from \Cref{fig:teaser}, following \cite{tang2023emergent}, we use features from the first upsampling block of the UNet from Stable Diffusion 2.1 with diffusion timestep $261$ as $f_\zeta$, as these features are similar for semantically-similar regions \cite{tang2023emergent,zhang2023tale,luo2024diffusion} but are locally smoother compared to DINO, which is consistent with the observations from \cite{zhang2023tale}.

\paragraph{Smoothness Loss.}
The smoothness loss from 
\cref{eq:forward_2d_to_2d} 
is specified as follows. 
Following \cite{ofri2023neural}, we define 
\begin{equation}
    \mathcal{L}_{\text{rigidity},\nabla}(T) := 
    \|J_\nabla(T)^T J_\nabla(T)\|_F + \|(J_\nabla(T)^TJ_\nabla(T))^{-1}\|_F,
\end{equation}
where $T$ jointly considers all neighboring coordinates $u$, instead of only one coordinate $u$ at once. We define $\left[T\right]_u = \tilde{u} - u$, where $\tilde{u}$ is the optimization variable for input $u$ from \cref{eq:forward_2d_to_2d}, and $J_\nabla(T)$ computes the Jacobian matrix of $T$ approximated with finite differences with pixel offset $\nabla$.
Following \cite{peebles2022gan}, we denote huber loss with $\mathcal{L}_\text{huber}$ and define the total variation loss as
\begin{equation}
    \mathcal{L}_\text{TV}(T) = \mathcal{L}_\text{huber}(\nabla_x T) + \mathcal{L}_\text{huber}(\nabla_y T),
\end{equation}
where $\nabla_x$ and $\nabla_y$ are partial derivatives \textit{w.r.t.} $x$ and $y$ coordinates, approximated with finite differences. 
The final smoothness loss is defined as 
\begin{equation}
    \mathcal{L}_\text{smooth} = 
    \lambda_{\text{rigidity},\nabla=10} + 
    0.1\mathcal{L}_{\text{rigidity},\nabla=1} + 
    10 \mathcal{L}_\text{TV}.
\end{equation}

\section{Feature Visualizations}
\label{app:analysis}
\begin{figure}[t]
    \centering
    \includegraphics[width=\linewidth]{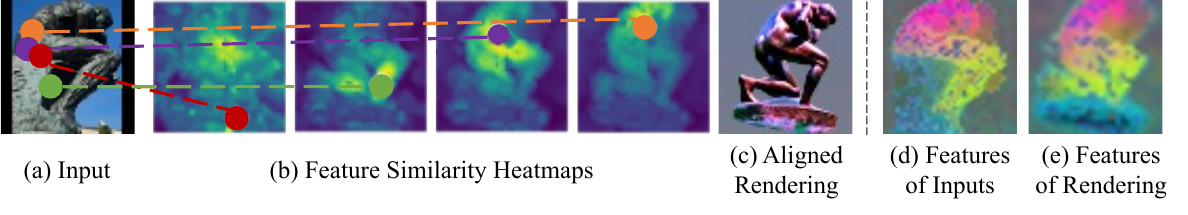}
    \vspace{-20pt}
    \caption{\textbf{Feature Visualizations.} Despite that DINO features tend to be noisy,
    our approach assigns a plausible pose to the input, as shown in the aligned rendering. 
    }
    \label{fig:exp_heatmap}
\end{figure}

Matching features independently for each pixel gives noisy similarity heatmaps (\Cref{fig:exp_heatmap} (b)), due to the noise of feature maps (\Cref{fig:exp_heatmap} (d-e)), and the lack of geometric reasoning in the matching process. 
Our method is robust to such noises as it seeks to align the input with a posed rendering considering all pixel locations in the input altogether. 

\section{Semantic Correspondence Matching}
\label{app:correspondence}

We provide additional quantitative evaluation of our method on the task of semantic correspondence matching. 
Given a pair of source and target image $(x_\text{source}, x_\text{target})$, and given a keypoint in the source image $u_\text{source}$, the goal of this task is to find its most semantically similar keypoint $u_\text{target}$ in the target image.

The matching process using our method is specified as follows. We first map the 2D keypoints being queried to the 3D coordinates in the canonical space, and then project these 3D coordinates to the 2D image space of the target image. 
Formally, given an image pair $(x_\text{source}, x_\text{target})$ and a 2D keypoint $u_\text{source}$, the corresponding keypoint $u_\text{target}$ is computed with
\begin{equation}
     u_\text{target} =
        \Phi_{x_\text{target}}^\text{rev} \circ \Phi_{x_\text{source}}^\text{fwd}(u_\text{source}),
\label{eq:source_to_target_to_source}
\end{equation}
with notations defined in \cref{subsect:optimization}.

For all experiments in this section, for 
\cref{eq:forward_2d_to_2d},
we set $\lambda_{\ell_2} = 10$ and for simplicity set $\lambda_\text{smooth} = 0$. 

\paragraph{Dataset.} 

We use SPair-71k~\cite{min2019spair}, a standard benchmark for semantic correspondence matching for evaluation. 
We evaluate our method on 9 rigid, non-cylindrical-symmetric categories from this dataset. 
The images for each category may contain a large diversity in object shape, texture, and environmental illumination. 

Following prior works~\cite{ofri2023neural,peebles2022gan}, we report the Percentage of Correct
Keypoints (PCK@$\alpha$) with $\alpha = 0.1$, a standard metric that evaluates the percentage of keypoints correctly transferred from the source image to the target image with a threshold $\alpha$. A predicted keypoint is correct if it lies within the radius of $\alpha\cdot \max(H_\text{bbox}, W_\text{bbox})$ of the ground truth keypoint in the object bounding box in the target image with size $H_\text{bbox}\times W_\text{bbox}$.

\begin{table}[t]
\centering
\begin{tabular}{lccccccccc|c}
\toprule
 & Aero & Bike & Boat  & Bus & Car & Chair & Motor & Train & TV & Mean \\\midrule
GANgealing.~\cite{peebles2022gan} & - & 37.5 & - & - & - & - & - & \\
Neural Congealing~\cite{ofri2023neural} & - & 29.1 & - & - & - & - & - & - & - & - \\
ASIC~\cite{gupta2023asic} &
57.9&
25.2&
24.7&
28.4&
30.9&
21.6&
26.2&
49.0&
24.6&
32.1\\
DINOv2-ViT-G/14~\cite{oquab2023dinov2} &
\textbf{72.5}&
67.0&
\textbf{45.5}&
54.6&
53.5&
40.7&
71.8&
\textbf{53.5}&
36.3&
55.0\\
Ours & 
70.0&
\textbf{70.3}&
40.0&
\textbf{65.8}&
\textbf{72.1}&
\textbf{50.1}&
\textbf{77.0}&
26.1&
\textbf{43.1}&
\textbf{57.2}
\\\bottomrule
\end{tabular}
\vspace{0.5em}
\caption{\textbf{Semantic Correspondence Evaluation on SPair-71k~\cite{min2019spair}.} Our method achieves an overall better keypoint transfer accuracy compared to prior 2D congealing methods and a 2D-matching baseline using the same semantic feature extractor as ours.}
\vspace{-1em}
\label{tab:spairs}
\end{table}

\begin{figure}[t]
    \centering
    \includegraphics[width=\linewidth]{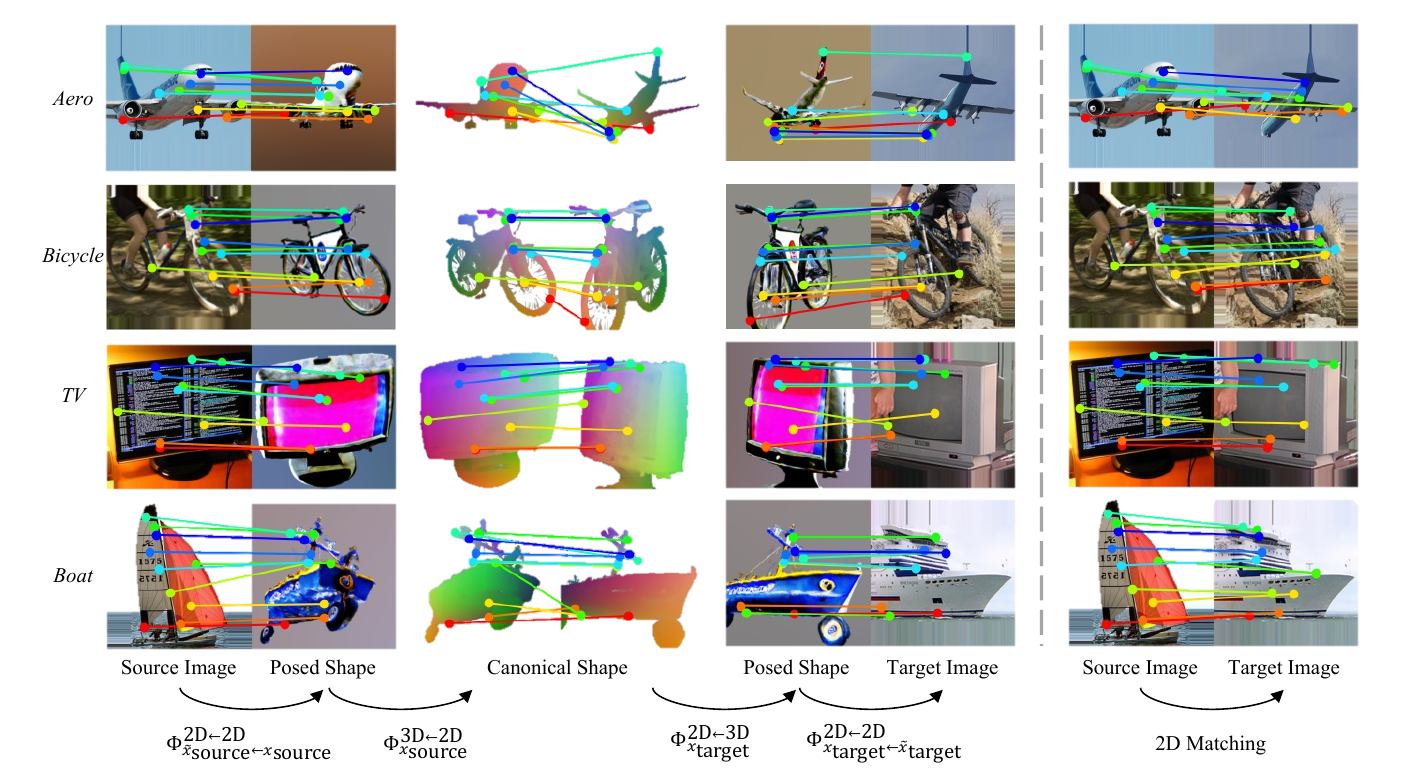}
    \caption{\textbf{Semantic Correspondence Matching.} 
    The figure shows results on 4 example categories from SPair-71k~\cite{min2019spair}. 
    To match a given keypoint from the source image, our method first warps the keypoint to the rendered image space (2D-to-2D), then identifies the warped coordinate's location in the canonical frame in 3D (2D-to-3D), then projects the \emph{same} 3D location to the rendering corresponding to the target image (3D-to-2D), and finally warps the obtained coordinate to the target image space (2D-to-2D).
    The learned 3D canonical shape serves as an intermediate representation that aligns the source and target images, and it better handles scenarios when the viewpoint changes significantly compared to matching features in 2D.
    }
    \label{fig:exp_spairs}
\end{figure}

\paragraph{Baselines.}
We compare with a 2D-correspondence matching baseline. Formally, for this baseline, for each querying keypoint $u_\text{query}$, we compute the keypoint prediction with
\begin{equation}
    u_\text{target} = \arg\min_u d_\zeta^{u, u} (x_\text{target}, x_\text{source}),
\end{equation}
where the distance metric $d_\zeta$ is defined in 
\Cref{eq:distance_pixelwise}
and is induced from a pre-trained features extractor $f_\zeta$. We use the same DINO feature extractor for our method and this baseline.

We further compare with previous congealing methods, GANgealing~\cite{peebles2022gan}, which uses pre-trained GAN for supervision, and Neural Congealing~\cite{ofri2023neural} and ASIC~\cite{gupta2023asic}, which are both self-supervised. 

\paragraph{Results.}
Results are shown in \cref{tab:spairs}. 
The performance gain over the DINOv2 baseline, which uses the same semantic feature extractor backbone as ours, suggests the effectiveness of 3D geometric consistency utilized by our framework.

Qualitative results are shown in \Cref{fig:exp_spairs}. Our method is the only one that performs correspondence matching via reasoning in 3D among all baselines. Such 3D reasoning offers an advantage especially when the relative rotation between the objects from the source and target image is large. Our method transforms the 3D coordinate from source to target in the canonical frame, where the 3D shape guarantees the 3D consistency. In comparison, as shown on the right of \Cref{fig:exp_spairs}, the baseline performs 2D matching and incorrectly matches the front of a plane with its rear, and incorrectly matches the front wheel of a bicycle with its back wheel.

\end{document}